# Dynamic Analytical Initialization Method for Spacecraft Attitude Estimators

Lubin Chang, *Member, IEEE,* and Fangjun Qin

*Abstract*—This paper proposes a dynamic analytical initialization method for spacecraft attitude estimators. In the proposed method, the desired attitude matrix is decomposed into two parts: one is the constant attitude matrix at the very start and the other encodes the attitude changes of the body frame from its initial state. The latter one can be calculated recursively using the gyroscope outputs and the constant attitude matrix can be determined using constructed vector observations at different time. Compared with traditional initialization methods, the proposed method does not necessitate the spacecraft being static or more than two non-collinear vector observations at the same time. Therefore, the proposed method can promote increased spacecraft autonomy by autonomous initialization of attitude estimators. The effectiveness and prospect of the proposed method in spacecraft attitude estimation applications have been validated through numerical simulations.

## I. INTRODUCTION

Determining the spacecraft attitude information is crucial for most space missions, which is achieved through the attitude determination and estimation algorithms. Here the attitude estimation and determination are stated to be different explicitly according to [1, 2]. Attitude determination refers to memoryless approaches

This paper is supported in part by the National Natural Science Foundation of China (61304241, 61374206 and 61503404) and the National Postdoctoral Program for Innovative Talents (BX 201600038).

The author is with the Department of Navigation Engineering, Naval University of Engineering, Wuhan 430033, China (e-mail: changlubin@163.com, Lubin.Chang.1987@ieee.org, fangjun_qin@163.com)

while attitude estimation algorithms can retain information from a series of measurements taken over time. Generally, attitude determination algorithms are based on solutions to Wahba's problem and attitude estimation algorithms are filtering based approaches. Compared with attitude determination, attitude estimation can make full used of the measurement information and can determine the dynamic attitude facilely. Moreover, attitude estimation can determine parameters other than the attitude, say gyro bias, resulting in a more accurate performance. In this respect, attitude estimation is more preferred in modern spacecraft missions.

The dynamic model of the spacecraft attitude estimation is virtually a nonlinear model, especially the vector observations based measurement model, which necessitates the nonlinear filtering algorithms. Among these filtering algorithms, the quaternion based extended Kalman filter referred as multiplicative extended Kalman filter (MEKF) is the most celebrated choice for the great majority of applications [3-7]. However, the MEKF may face difficulty when the dynamical models are highly nonlinear or/and a good a priori state information can not be obtained, which promotes the development of advanced nonlinear attitude estimators [8-14]. The performance improvement of these advanced nonlinear attitude estimators is at the cost of increasing complexity and computational burden. Meanwhile, these estimators may be very difficult to tune and their stability has not been proven. Taking the unscented quaternion estimator (USQUE) for example [8], the initial attitude covariance setting has a great impact on the filtering performance, which is relative to the initial attitude estimate error that we can not know. A conservative method is to set the attitude covariance to be very large, which may result in a very slowly convergent speed and large steady-state error.

For the special spacecraft attitude estimation problem using vector observations, the corresponding nonlinearities of the model are determinate. In this respect, the superiority of the advanced nonlinear attitude estimators over MEKF lies only on their capacity to handle the large initial estimate error. If the initial attitude information can be obtained to certain precision, the MEKF is still the most preferred choice for real-time spacecraft attitude estimation. These facts represent the main motivation of this paper, which is devoted to proposing a novel initialization method for the MEKF. The advanced nonlinear attitude estimators are not preferred to initialize the MEKF, since their covariance initialization is still a cumbersome problem. The traditional attitude determination methods can not handle the dynamic attitude estimation owing to their memoryless characteristic. In the proposed initializing method, the spacecraft attitude is decomposed into two parts based on introducing a new inertial fixed frame. The first one encodes the attitude changes of the body frame from its initial state, which can be derived through attitude calculation using the gyro measurements with the naturally known initial value. The other one encodes the constant attitude between the body frame and inertial frame at the very start of one mission. This constant attitude can be determined based on the constructed new vector observations. Through such attitude decomposition, the heart of the attitude estimation has been transformed into determining the constant attitude using vector observations at different time. That is to say, the attitude determination methods can be used to determine the dynamic attitude using the series of measurements taken over time. The proposed procedure makes the attitude determination methods be with memory. The resulting attitude determination results within a short time period are accurate enough to guarantee the validity of the linearization in the MEKF. Therefore, autonomous initialization of attitude estimators can be expected by the proposed methodology.

The attitude decomposition method is enlightened by our previously studied initial alignment methods for Strapdown inertial navigation system [15-19]. Hopefully, the investigated method can provide a new way of thought for the spacecraft attitude estimators initialization problem.

## II. SPACECRAFT ATTITUDE ESTIMATION MODEL

The discrete-time attitude kinematics model is given by [2]

$$q_{i,k}^b = \Omega(\omega_{k-1}) q_{i,k-1}^b \tag{1}$$

with

$$\Omega(\omega_{k-1}) = \begin{bmatrix} Z_{k-1} & \varphi_{k-1} \\ -\varphi_{k-1}^T & \cos(0.5\|\omega_{k-1}\|\Delta t) \end{bmatrix}$$

$$Z_{k-1} = \cos(0.5\|\omega_{k-1}\|\Delta t) I_{3\times 3} - (\varphi_{k-1} \times)$$

$$\varphi_{k-1} = \sin(0.5\|\omega_{k-1}\|\Delta t)\omega_{k-1}/\|\omega_{k-1}\|$$

where $q_{i,k}^b$ denotes the attitude quaternion from the inertial frame $i$ to the body frame $b$. $\Delta t$ is the sampling interval in the gyro. $I_{3\times 3}$ is the identity matrix with denoting dimension. $\omega_{k-1}$ is the angular rate of the spacecraft and can be derived from the gyro measurement as

$$\omega_{k-1} = \tilde{\omega}_{k-1} - \beta_{k-1} - \eta_{v,k-1} \tag{2}$$

where $\tilde{\omega}_{k-1}$ is the measurement of the gyro. $\eta_{v,k-1}$ is the Gaussian white noise process. $\beta_{k-1}$ is the gyro bias and is assumed to be constant, that is

$$\beta_k = \beta_{k-1} \tag{3}$$

The vector observation model for attitude estimation is given by

$$b_k = A(q_{i,k}^b)r_k + n_k \tag{4}$$

where $b_k$ is the body-frame vector and $r_k$ is the reference-frame vector. $A(q_{i,k}^b)$ is the direction cosine matrix or rotation matrix corresponding to the attitude quaternion $q_{i,k}^b$. $n_k$ is the Gaussian white measurement noise.

Eq. (1), (3) and (4) constitute the dynamic model for spacecraft attitude estimation with (1) and (3) being the process model and (4) being the measurement model.

### III. NOVEL INITIALIZATION METHOD

Generally, at the beginning of one mission, we can not obtain the precise initial attitude and gyro bias. So the attitude should be calculated recursively using (1) with only a guess of the initial value. The corresponding calculation error can be estimated based on the vector observation model (4) and used to refine the calculated attitude. If the gyro bias is not considered provisionally, the attitude error during its recursive calculation is mainly caused by its initial error. With this consideration, we introduce a new inertial frame $b_0$ by fixing the body frame $b$ at the start-up in the inertial space. Then the attitude matrix can be decomposed into two parts as

$$A(q_i^b) = A(q_{b_0}^b) A(q_i^{b_0}) \tag{5}$$

It is clearly that $A(q_i^{b_0})$ is a constant matrix. The attitude kinematics model for $q_{b_0}^b$ is given by

$$q_{b_0,k}^b = \Omega(\omega_{k-1}) q_{b_0,k-1}^b \tag{6}$$

It is shown that the input angular rate for calculation of $q_{b_0,k}^b$ is the same with that for $q_{i,k}^b$ and is provided by the gyro. This is because that $b_0$ and $i$ are both inertial frame and the gyro just measures the body angular rate relative to the inertial frame. According to the definition of the inertial frame $b_0$, it can be easily obtained that the initial value for (6) is given by

$$q_{b_0,0}^b = q_{b_0}^{b_0} = [0;0;0;1]$$

That is to say, $q_{b_0,k}^b$ can be calculated recursively by (6) using the gyro measurements without any initial error.

After the aforementioned attitude matrix decomposition, the heart of estimation of $q_{i,k}^b$ has been transformed into determination of the constant attitude $q_i^{b_0}$. Next, we will present the determination method for $q_i^{b_0}$ based on the vector observations.

Substituting the attitude decomposition result (5) into (4) yields (With no consideration of the measurement noise provisionally)

$$b_k = A(q_{b_0}^b) A(q_i^{b_0}) r_k \tag{7}$$

Multiplying $A(q_b^{b_0})$ on both sides yields

$$A(q_b^{b_0}) b_k = A(q_i^{b_0}) r_k \tag{8}$$

Given the gyro measurements, $q_{b_0,k}^b$ can be calculated recursively and therefore, $A(q_{b,k}^{b_0})$ can be viewed as a known quantity. Denote $\bar{b}_k = A(q_{b,k}^{b_0}) b_k$, (8) can be rewritten as

$$\bar{b}_k = A(q_i^{b_0}) r_k \tag{9}$$

Eq. (9) is a typical attitude determination problem using vector observations and its Wahba's problem formulation can be given by

$$J(A) = \frac{1}{2}\sum_{k=1}^{M}\left\|\bar{b}_k - A(q_i^{b_0})r_k\right\|^2 \qquad (10)$$

where $M$ is the number of used vector observations in the initialization process. Many existing algorithms can be used directly to address this problem [11], such as the Davenport's q method used in this paper.

After attitude $q_i^{b_0}$ has been determined, the spacecraft attitude can be readily obtained through (5).

Determine the following two matrices

$$\bar{b}_k^+ = \begin{bmatrix} 0 & -\bar{b}_k^T \\ \bar{b}_k & (\bar{b}_k \times) \end{bmatrix}, r_k^- = \begin{bmatrix} 0 & -r_k^T \\ \bar{b}_k & -(r_k \times) \end{bmatrix} \qquad (11)$$

The explicit procedure of the proposed initialization method is summarized in **Algorithm 1**.

**ALGORITHM 1**: THE PROPOSED INITIALIZATION METHOD

---

*Initialization*: Set $k = 0$. Let $q_{b_0,0}^b = q_{b_0}^{b_0} = [0;0;0;1]$ and $K_0 = 0_{4\times 4}$.

*Step 1*: Set $k = k+1$.

*Step 2*: Update $q_{b_0,k-1}^b$ to $q_{b_0,k}^b$ according to (6).

*Step 3*: Construct vector observations $\bar{b}_k = A(q_{b,k}^{b_0})b_k$.

*Step 4*: Construct $\bar{b}_k^+$ and $r_k^-$ according to (11).

*Step 5*: Update $K_{k-1}$ to $K_k$ according to

$$K_k = K_{k-1} + \left[(\bar{b}_k^+ - r_k^-)^T(\bar{b}_k^+ - r_k^-)\right]\Delta t.$$

*Step 6*: Determine $q_i^{b_0}$ by calculating the normalized

eigenvector of $K_k$ belonging to the smallest eigenvalue.

*Step 7*: Obtain the attitude matrix at current time

$$A(q_{i,k}^b) = A(q_{b_0,k}^b)A(q_i^{b_0}).$$

*Step 8*: Go to *Step 1* until the end of the initialization period.

---

**Remark 1**: The investigated problem is the dynamic spacecraft attitude estimation and the heart of the solution strategy is the constant attitude determination. Through the proposed transformation, the resulting attitude determination seem to be with memory as the vector observations at different time instant are all used to determine the same attitude. With such property of being memory, a very fast convergent speed can be expected for the transformed attitude determination problem. Therefore, after a short time, the determined attitude will guarantee the validity of the linearization in the following applied MEKF. That is to say, the proposed initialization method is very efficient.

**Remark 2**: Actually, the proposed method can also be used all through the mission, not only as an initialization method for the attitude estimators. However, this is not recommended. It is known that the proposed method is virtually an analytical method and only the attitude can be determined. The gyro bias, as an another main error source, can not be estimated and compensated in the proposed method. Meanwhile, the noise inherent in the vector observations expressed in the body frame can also be not handled by the proposed method. Therefore, if the proposed method is used all through the mission, the resulting attitude precision may be not so satisfactory. Actually, the gyro bias is the main error source for the proposed method and it will be cumulated into the determined attitude and therefore, there will be a slowly climbing trend in the attitude determination result. This is just another reason why the proposed method is used within a short time period at the start of the mission.

**Remark 3**: If the spacecraft is static or there are more than two non-collinear vector observations at the same time, the attitude determination methods can also be used to initialize the attitude estimators. However, the aforementioned requirements restrict the autonomous initialization under arbitrary conditions. In contrast, the proposed method can be effective even when the spacecraft is dynamic or there is only one vector observation at one time. In this respect, the proposed method is more versatile for the initialization of the attitude estimators.

## IV. SIMULATION EXAMPLE

In this section, the performance of the proposed attitude estimation method is evaluated through several test cases using the simulation example 7.2 of [1] (also the example 6.2 of [2]). A 90-min simulation run is shown. The simulation here is a little different with that in [1]. In [1], the star tracker can sense up to 10 stars, while in this simulation example only one star's measurement is used in each time instant. This makes the problem more difficult to address using conventional attitude determination methods. In the proposed attitude estimation methodology, denoted as "Optimal+MEKF", the Davenport's q method is applied for the first 5 minutes using our constructed observations, followed by the MEKF for the remaining time. The well-known attitude estimators, i.e. MEKF and USQUE are also evaluated for comparison. Moreover, the method that making use of only the Davenport's q method based on our constructed observations is also evaluated and it is denoted as "Optimal" here.

In the first case, the initial attitude estimate error is set as $[10\ 10\ 30]$ deg. Actually, this attitude estimate error setting is only for the MEKF and USQUE. For the "Optimal+MEKF" methodology, any prior information is meaningless. The gyro bias is set to 0.1 deg/h for each axis and the initial bias estimate is set to 0 for each axis. The initial covariance for the attitude error is set to $(0.1 \deg)^2$ for the "Optimal+MEKF". As is known, this initial covariance is firstly used after the 5 minutes' Davenport's q method being performed. The initial covariance for the attitude error is set to $(10 \deg)^2$ for both the MEKF and USQUE. The initial covariance for the gyro bias is set to $(0.1 \deg/h)^2$ for "Optimal+MEKF", MEKF and USQUE. 50 Monte Carlo runs were carried out and the norm of total attitude estimation error for this case is shown in Fig. 1. It is known that the validity of the linearization in the MEKF relies heavily on the roughly known initial attitude estimate. In this case, the initial attitude estimate error is too large to guarantee such validity and the resulting filtering performance is much degraded as shown in Fig. 1. The USQUE can handle such large initial attitude estimate error and the resulted filtering performance is much more accurate. However, the performance improvement is at the cost of large computational burden. In our simulation, the computational cost of USQUE is almost 5 times of that of the MEKF. For the "Optimal+MEKF", after 5 minutes' attitude determination procedure, the attitude error can be reduced to 0.08 deg which is small enough for the validity of the linearization in the MEKF. So the following MEKF achieves a very accurate estimation result, nearly the same as the USQUE but with much less computation burden. It is shown that there is a slowly climbing trend in the attitude determination result by "Optimal", which is owed to the existence of gyro bias that has not been taken into account in the algorithm. It is also shown that the attitude determination error can be reduced to within 1 degree almost instantaneously. Such fast convergent speed and accurate performance make this method very suitable for initializing the attitude estimator, more specifically, the MEKF.

In the second case, the initial attitude estimate error is set as $[30\ 30\ 60]$ deg. The setting for the gyro bias is same as that in the first case. This different attitude estimate error setting with the first case can only affect the performance of the MEKF and USQUE. The initial covariance for the attitude error is set to $(25\deg)^2$ for both the MEKF and USQUE. The parameters settings for "Optimal+MEKF" are all the same with that in the first case. The averaged norm of total attitude estimation error over 50 Monte Carlo runs for this case is shown in Fig. 2. In this case, the performance of the MEKF has been much degraded compared with the first case. The performance of the USQUE has also been degraded a little. In this case, the "Optimal+MEKF" methodology has outperformed the USQUE with much less computational burden. It can be deduced that the performance of the USQUE can be further degraded when the initial attitude estimate error becomes larger. That is to say the USQUE can not handle arbitrary large initial attitude estimate error. In contrast, the proposed "Optimal+MEKF" can be carried out with nothing prior attitude estimate information. So the "Optimal+MEKF" will be more celebrated in realistic application.

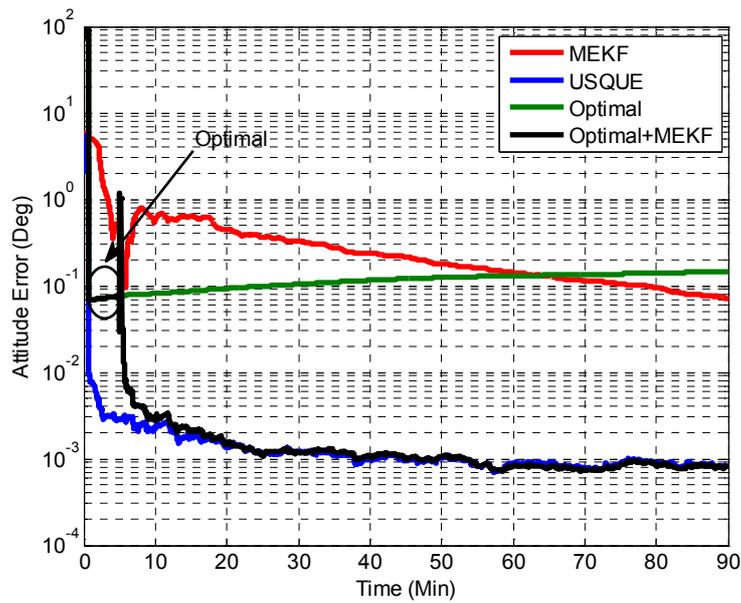

Fig. 1. Attitude estimate errors by different methods (case 1)

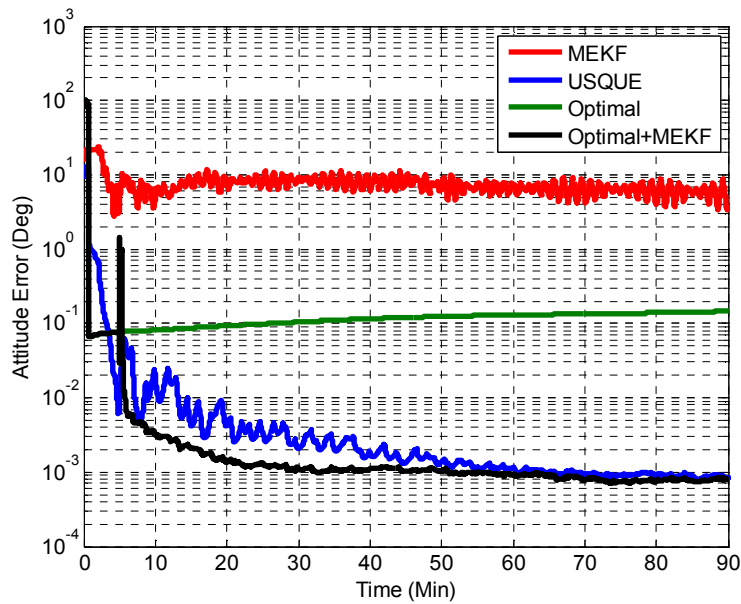

Fig. 2. Attitude estimate errors by different methods (case 2)

It has been pointed out that the performance of the constructed attitude determination method relies heavily on the precision of the gyro bias. In the last two cases, the gyro bias is set to 0.1 deg/h for each axis, which can be viewed as a very accurate level. The resulting attitude determination performance is therefore very accurate, as shown in Fig. 1 and 2. In the third case, the gyro bias is set to 10 deg/h for each axis and the initial bias estimate is set to 0 for each axis. The initial covariance for the gyro bias is set to $(10\deg/h)^2$ for "Op-

timal+MEKF", MEKF and USQUE. The initial attitude estimate error is set as $[10\ 10\ 30]$ deg and the other settings are all the same with that in the first case. The averaged norm of total attitude estimation error over 50 Monte Carlo runs for this case is shown in Fig. 3. For the "Optimal+MEKF", after 5 minutes' attitude determination procedure, the attitude error is reduced to 2.95 deg which is much larger than that in the last two cases. This is because that in this case a much low grade gyro has been used. However, the attitude determination error by the proposed method is also small enough for the validity of the linearization in the MEKF as the following MEKF still has an accurate estimate performance. The performance of the MEKF and USQUE is similar with the corresponding one in the first case, which indicates that the gyro bias can be well estimated by the attitude estimators, no matter how large it is.

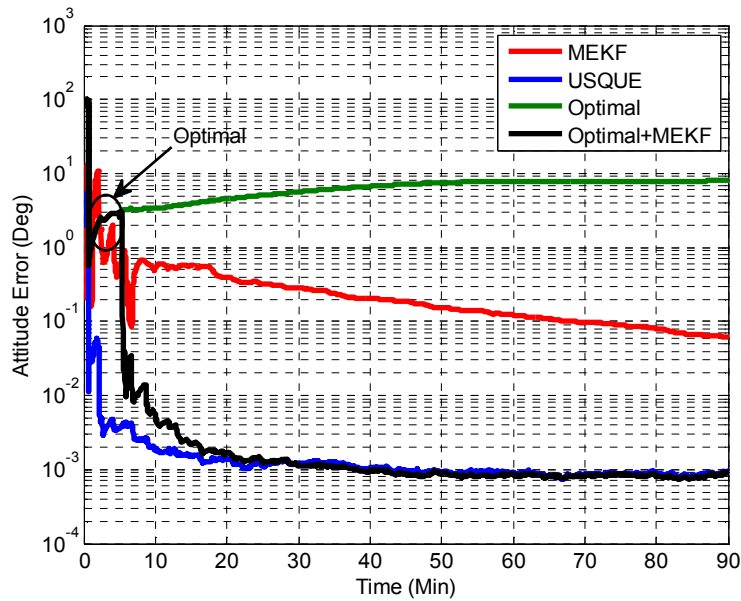

Fig. 3. Attitude estimate errors by different methods (case 3)

In the fourth case, the initial attitude estimate error is set as $[30\ 30\ 60]$ deg and the corresponding initial covariance for the attitude error is set to $(25\deg)^2$ for both the MEKF and USQUE. The other settings are all the same with that in the third case. The averaged norm of total attitude estimation error over 50 Monte Carlo runs for this case is shown in Fig. 4. It is shown that the performance of both the MEKF and USQUE has been much degraded. The superiority of "Optimal+MEKF" over USQUE becomes more obvious.

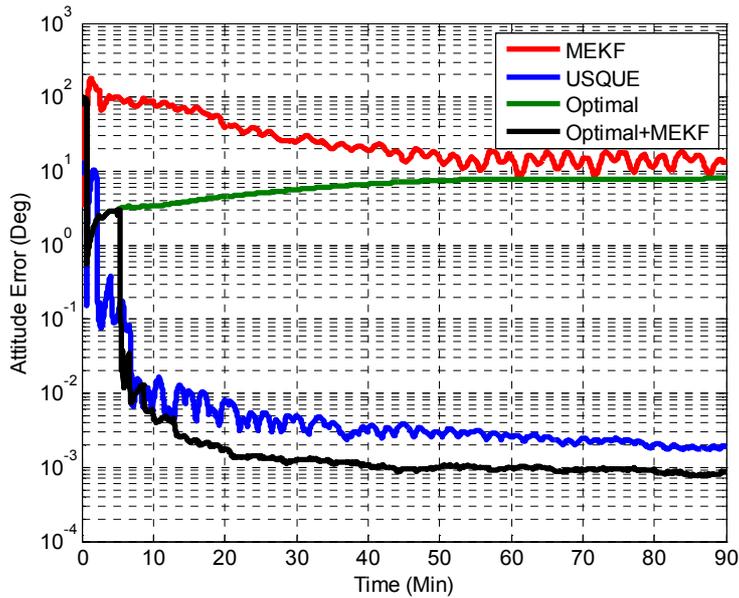

Fig. 4. Attitude estimate errors by different methods (case 4)

Since the degree of the gyro bias has a significant effect on the performance of the proposed method, the performance of the proposed method with different degrees of the gyro bias has also been evaluated. The corresponding attitude determination error is shown in Fig. 5. It is clearly shown that when the gyro bias becomes larger, the determined attitude is degraded correspondingly. Specifically, when the degree of the gyro bias exceeds 10deg/h, the proposed method is not recommended for initializing the MEKF. For such cases however, it can be used to initialize the advanced nonlinear attitude estimators, such as the USQUE. From Fig.5, it can also be seen that even when the gyro bias is small enough, such as 0.01deg/h, the determined attitude error is still larger than that of the MEKF with appropriate initialization. The reasons have been discussed in **Remark 2**, that is, the proposed method is virtually an analytical method and the noise inherent in the vector observations can not be well handled. In this respect, the proposed method is not recommended as the attitude determination method through the whole mission even when the gyro bias is very small.

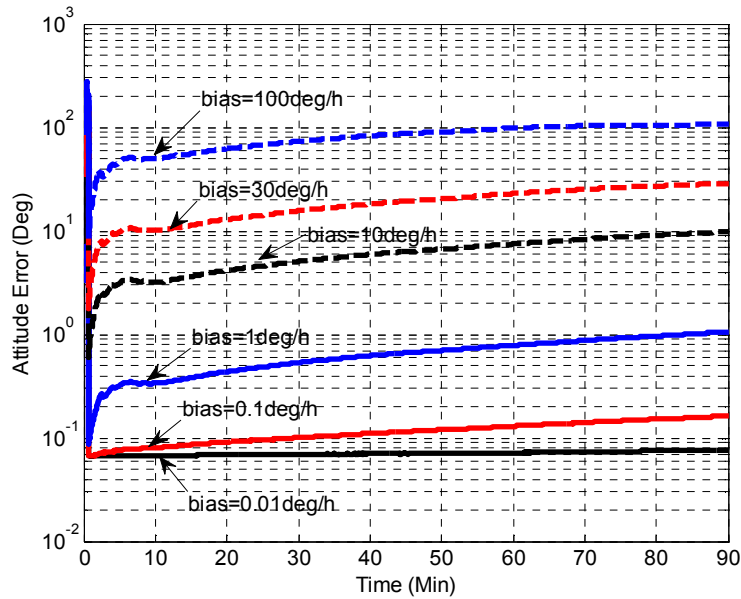

Fig. 5. Attitude estimate errors by "Optimal" with different gyro bias

## V. Conclusions

In this paper, a novel initialization method is proposed for spacecraft attitude estimators. In the proposed method, the attitude is decomposed into two parts: one encodes the attitude changes of the body frame and the other encodes the constant attitude between the body frame and inertial frame at the very start of one mission. The constant attitude can be determined using the constructed vector observations at different time, which makes the proposed attitude determination method be with memory. Simulation results indicated that the proposed method can determine the attitude to a quite precise degree within only a short time period. With the initialized value provided by the proposed method, the MEKF can estimate the attitude quite accurate. By the proposed initialized method, some complex nonlinear attitude estimators are no longer needed. Since the proposed method needs nothing prior information, autonomous initialization of attitude estimators can be expected.

## Acknowledgment

The authors would like to thank Prof. Crassidis and Junkins for providing the MATLAB programs for the examples of Ref. [1] in *http://www.buffalo.edu/~johnc/estim_book.htm*. These programs have given the au-thors a more thoroughly understanding of the attitude estimation algorithms and accelerated the corresponding investigation in this paper.